# Real-time Robot-assisted Ergonomics*

A. Shafti, A. Ataka, B. Urbistondo Lazpita, A. Shiva, H.A. Wurdemann and K. Althoefer.

*Abstract*— This paper describes a novel approach in human-robot interaction driven by ergonomics. With a clear focus on optimising ergonomics, the approach proposed here continuously observes a human user's posture and by invoking appropriate cooperative robot movements, the user's posture is, whenever required, brought back to an ergonomic optimum. Effectively, the new protocol optimises the human-robot relative position and orientation as a function of human ergonomics. An RGB-D camera is used to calculate and monitor human joint angles in real-time and to determine the current ergonomics state. A total of 6 main causes of low ergonomic states are identified, leading to 6 universal robot responses to allow the human to return to an optimal ergonomics state. The algorithmic framework identifies these 6 causes and controls the cooperating robot to always adapt the environment (e.g. change the pose of the workpiece) in a way that is ergonomically most comfortable for the interacting user. Hence, human-robot interaction is continuously re-evaluated optimizing ergonomics states. The approach is validated through an experimental study, based on established ergonomic methods and their adaptation for real-time application. The study confirms improved ergonomics using the new approach.

## I. Introduction

Collaborative robots or so-called cobots are opening new possibilities in human-robot interaction within industrial environments. Major design factors in the creation of collaborative robots are health and safety. The main issues covered within the area of health and safety in human-robot interaction are typically those relating to collision avoidance and ensuring that the human user is safe from immediate injury in case the robot and/or the user are not within their anticipated trajectory as well as behaviour due to any sort of failure or error. This is indeed critically important, and avoiding these states leads to averting immediate harm from the human. However, there is less emphasis on considering the human's long-term health and safety. Issues relating to the worker's comfort during working hours on the factory floor relate directly to their long-term health. Work related musculoskeletal disorders (WMSDs) are the result of a workers' comfort issues going unnoticed for a prolonged period of time. WMSDs are not only an issue of personal health for the worker, they also affect the business interests of the company they are working for. Prevention of WMSDs is therefore critical and of high importance.

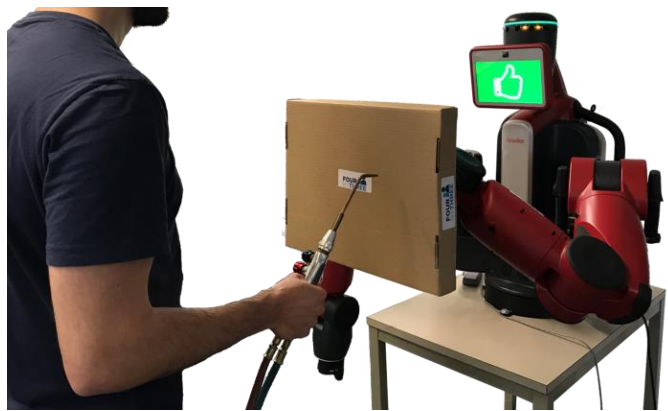

**Figure 1** – Robot-assisted ergonomics: the robot monitors the human's ergonomic state in real-time, understands it and optimises the relative pose between the human and the workpiece constantly improving ergonomics.

Industrial workplaces are changing. The advances in safe human-robot interaction (HRI) have led to industrial robots moving past the large, heavy, fenced robots working on their own and towards relatively small, lightweight and safe robots that work hand-in-hand with human users [1]. This presents an opportunity to considerably improve ergonomics and comfort within the industrial workplaces through automation and with real-time ergonomics monitoring and response through robot assistance. This paper presents a novel interaction approach, where the robot can sense the human user's ergonomic state based on established posture-monitoring methods, such as RULA, and is thus able to react to it, with the aim to constantly improve the human's ergonomic state – in effect, optimising the interactions based entirely on the human's comfort and ergonomics. The RULA worksheet by ErgonomicsPlus® is presented here in Figure 2 for reference throughout the paper.

There is robotics research where ergonomics methods have been considered. In [2], the concept of ergonomics-for-one, i.e. the fitting of task and tool design to a specific person with special needs, is used for HRI. The authors create a robotic shopping cart for visually impaired users basing their design decisions on individual interviews with the users. The work in [3] concerns the creation of a nine degree-of-freedom model of the human arm to be used in the development, testing and comfort optimisation of an exoskeleton for the upper-arm. The authors report that the use of this method has resulted in the device being able to interact more comfortably with the human, and with more use of the natural limb workspace leading to better integration with human movements. In [4], a humanoid robot's motion and manipulation planning is based on the RULA directives for

* This work was supported by the Horizon 2020 Research and Innovation Program under grant agreement 637095 in the framework of EU project FourByThree.

A. Shafti is with the Dept. of Computing, Imperial College London. A. Ataka, B. Urbistondo Lazpita and A. Shiva, are with the Centre for Robotics Research at King's College London.

H. A. Wurdemann is with the Department of Mechanical Engineering, University College London, Gower St., London WC1E 6BT, UK – h.wurdemann@ucl.ac.uk.

K. Althoefer is with the School of Engineering and Materials Science, Queen Mary University of London, E1 4NS, London, UK – k.althoefer@qmul.ac.uk.

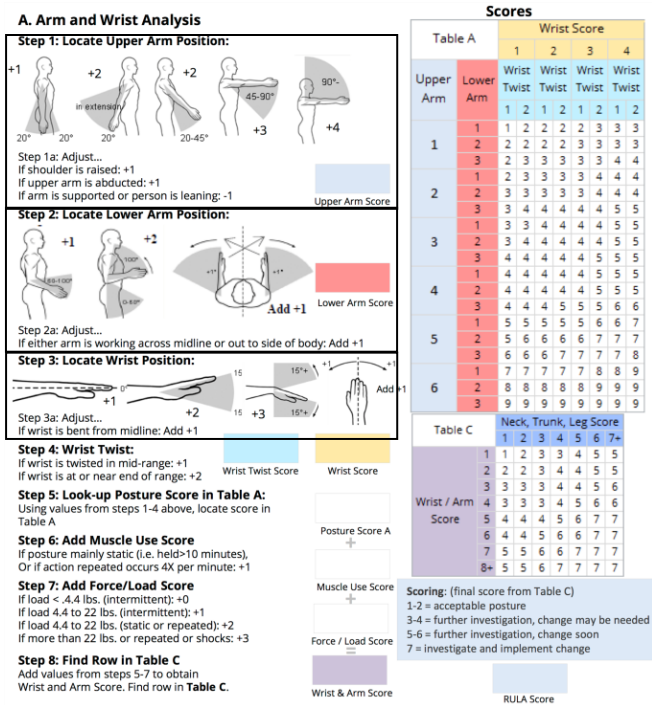

**Figure 2** – The Rapid Upper Limb Assessment (RULA) worksheet (arm/wrist analysis only). Step 1 (upper arm), Step 2 (lower arm) and Step 3 (wrist) are highlighted. From the user's joint angles, the ergonomics level is determined.

human comfort, leading to human-like movements by the robot, aiming to achieve improved interaction with humans as the robot's human-like movements will be more easily understood by the interacting human. In [5], a scooter robot learns the paths that the worker tends to use, and adjusts its wheel axes directions in a manner to reduce the forces applied by the user, improving the ergonomics of the task. In [6], the use of software-ergonomics to create more ergonomic collaborative tasks with the use of robots is proposed.

Recently, there has been an increase in efforts to bring ergonomic methods into the realm of HRI [7-11]. Here we present a computationally light method for human robot interactions based only and entirely on the optimization of the human user's ergonomic state. In this manner, the type of handover industrial assistance cases described here will not need any other type of planning or programming, but rather, the system will just attempt to make the human comfortable, which we propose leads to optimal fulfilment of the task at hand.

## II. METHODS

### A. Sensor-based, real-time ergonomics assessment

In our set-up, we use a suite of sensors (including the Kinect depth sensor and inertial measurement units) to determine the posture of the interacting human and the Baxter® Research Robot to adjust the interaction objects so that ergonomics can be achieved. Our algorithms run in the Robot Operating System (ROS). In short, we aim to achieve robot assisted ergonomics integrating the posture ergonomics assessment method (here, Rapid Upper Limb Assessment - RULA) into our system.

The Kinect™ (which had been previously used with the RULA method in [12] and other ergonomic techniques in [13], [14]) is employed to 'see' the user's posture as a human inspector would do [15]. Using open source libraries to use with the Kinect™ a human's skeleton frames are broadcast, providing 15 joint positions, among them the head, neck, torso, shoulder, elbow and hand, that can be represented in a vectorized form (Figure 3 (left)).

As Figure 2 shows a number of RULA specific angles related to the upper-arm, lower-arm and wrist are required to be computed from the Kinect™ output and to projected into the human body's anatomically defined planes, (Figure 4).

Any deviations from the ergonomic optimum caused by these angles can be referred to as deviations in the corresponding planes, e.g. 'lower-arm is in transversal deviation'. These angles are illustrated on a human body model in Figure 5.

To calculate the upper-arm and lower-arm angles, the dot product of limb vectors is used, as follows. For the lower-arm's sagittal angle, the position data of the following joints is used: the hand (wrist), elbow, and shoulder. The positions for these are given as $r_h$, $r_e$ and $r_s$ respectively. All positions are measured with respect to the shoulder frame. Vectors are created by subtracting $r_e$ from $r_h$ and $r_s$ from $r_e$. Then, the lower-arm sagittal angle, formed between the hand-elbow and elbow-shoulder vectors as described in Figure 5, is given by:

$$\beta_s = \cos^{-1}\left[\frac{(r_h - r_e)^T (r_e - r_s)}{|r_h - r_e| \cdot |r_e - r_s|}\right]$$

The upper-arm's sagittal angle, based on the position data of the following joints, the torso ($r_t$), neck ($r_n$), elbow ($r_e$) and shoulder ($r_s$), can be computed as follows:

$$\alpha_s = \cos^{-1}\left[\frac{(r_t - r_n)^T (r_e - r_s)}{|r_t - r_n| \cdot |r_e - r_s|}\right]$$

Using the position data of the following joints, neck ($r_n$), elbow ($r_e$) and shoulder ($r_s$) measured with respect to the shoulder frame, the upper-arm's coronal angle can be calculated as follows:

$$\alpha_c = \cos^{-1}\left[\frac{(r_n - r_s)^T (r_e - r_s)}{|r_n - r_s| \cdot |r_e - r_s|}\right] - 90^0$$

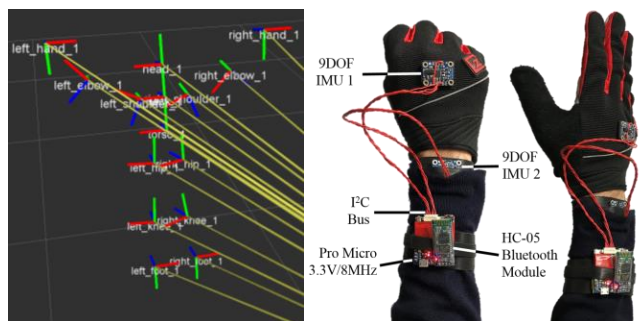

**Figure 3** – (left) The human skeleton as detected by the Kinect™ camera and visualised in RVIZ; (right) The wearable double Inertial Measurement Unit (IMU) setup used to detect wrist angles in real-time.

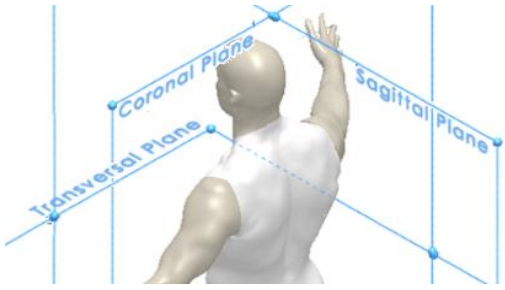

**Figure 4** – Anatomical planes of the human body. The *sagittal* plane divides the body into a left and right side. The *coronal* plane divides it into a front and rear side. The *transversal* plane divides it into an upper and lower side.

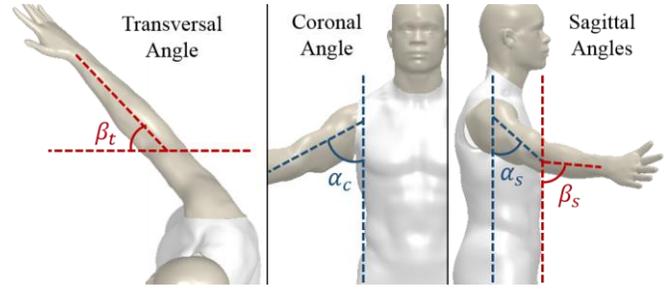

**Figure 5** – View of the human body, showing ergonomic deviation angles projected on the *sagittal*, *coronal* and *transversal* planes of the body.

Finally, the lower-arm's transversal angle based on the position data of the following joints, the neck ($r_n$), hand ($r_h$), elbow ($r_e$) and shoulder ($r_s$) can be calculated as follows:

$$\beta_t = 90^0 - \cos^{-1}\left[\frac{(r_s - r_n)^T (r_h - r_e)}{|r_s - r_n| \cdot |r_h - r_e|}\right]$$

The above provides all the upper-arm and lower-arm angles required to fulfil steps 1 and 2 of the RULA worksheet as shown in Figure 2.

The above angles suffice to detect the type of ergonomic deviation. To avoid errors in the detection of thresholds due to shortcomings of the Kinect™, particularly within the non-sagittal movements possibly due to issues within the OpenNI library, flags are defined as a redundancy to ensure the type and direction of deviation are detected correctly [7].

The data received from the Kinect™ camera is not enough to make judgements on wrist angles. Here we use two Inertial Measurement Units (IMUs) Bosch® BNO055 Absolute Orientation Sensor. These two BNO055 sensors, each providing 9 degrees of freedom (DOF) data; 3-axis acceleration, 3-axis gyroscope and 3-axis magnetometer data, are integrated with a microcontroller and auxiliary electronics and embedded in a glove and wristband, respectively [7]. Two of the BNO055 sensors are used, one of them to be placed on the lower-arm as reference, and the other to be placed on the back of the hand to determine the wrist bending and twist angles (steps 3 and 4 of RULA worksheet, table A in Figure 2).

Figure 3 (right)) shows the IMU sensors setup, as well as the printed circuit board (PCB) created to interface the sensors. Using the Kinect™ and the IMU sensors, all necessary angles to calculate the RULA score for the human arm, Figure 2, can be determined in real time.

*B. Non-ergonomic state avoidance through robot assistance*

The next step is to use the RULA score calculated based on the acquired sensor information, to decide on the suitable robot response to improve ergonomics. Looking at the full RULA arm score, 144 different ergonomic states exist. However, considering that the full RULA score for the arm is based on individual scores for the different arm sections (i.e. steps 1-4 in Figure 2), it is possible to determine which section of the arm is in a non-ergonomic state by monitoring the individual scores. Hence, if individual robot responses are created for each section of the arm for an ergonomic optimum, one can respond to all ergonomic states, as the appropriate overall response will always be a combination of the individual responses.

The arm is divided, within RULA, to the "upper-arm", "lower-arm" and "wrist". Each of these have two angles being monitored. The upper-arm's sagittal and coronal, lower-arm's sagittal and transversal angles and the bend and twist angles for the wrist (refer to Figure 5). Therefore, any particular RULA arm score can be broken down into scores for the upper-arm, lower-arm and wrist, each of which could be outside the ergonomic optimum area defined for it. Once the source of non-ergonomics is identified, the response will be to move that particular section of the arm back into its ergonomic optimum.

- *Upper-arm responses*

Looking at the RULA description for the upper-arm (Figure 6) the anatomic figures show the scores for different ranges of sagittal angles. These are scored from 1 to 4. Step 1 also provides adjustments to the score, the relevant one of which is "if upper-arm is abducted add +1". This is referring to the upper-arm coronal angle and means that if this projected value has any value larger than 0 (which in practice would mean an abduction of the upper-arm) then the upper-arm score should be increased by one point. In our algorithm, this threshold is set to 10⁰ to avoid unwanted reactions for small deviations.

The upper-arm being in a non-ergonomic region based on its sagittal angle can occur in two scenarios: (i) upper-arm in angles larger than 20⁰ to the front (Figure 6). The reason for this would be the workpiece being higher than the human user's ergonomic zone and also out of their reach. Therefore, the response should be for the workpiece to be moved down and towards the user. (ii) upper-arm in angles larger than 20⁰ to the back. The reason for this would be the workpiece being too close to the human user. Therefore, the response should be for it to be moved further away from them.

The responses for the upper-arm are therefore in the form of a translation. The human is trying to reach a particular target zone on the workpiece which is outside their ergonomic zone. For the human's arm to move back into the ergonomic optimum, the target zone needs to be moved into their ergonomic reach. The robot tool centre point's (TCP) response translation will be the same as the translation that would take the human's arm back into the ergonomic position. Effectively, the human will be following the targeted area on the workpiece as the robot moves it into a more ergonomically reachable position. The robot's translation values therefore depend on how far the human arm is from its ergonomic position. As the sensing system provides actual deviation angles, it is possible to find the values for these translations, Figure 6.

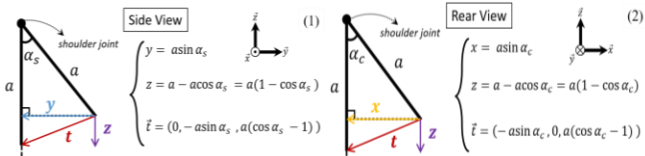

**Figure 6** – (1) Calculations for the translation to improve the upper-arm's deviation in the *sagittal* plane. (2) Calculations for the translation to improve the upper-arm's deviation in the *coronal* plane. The value 'a' refers to the human's shoulder-elbow length.

The line marked as 'a' represents the human upper-arm, viewed from the side, with 'a' being the shoulder-elbow length and $α_s$ the angle by which the upper-arm is deviated in the sagittal plane. Looking at the RULA descriptions for the upper-arm in Figure 6, it is obvious that the most ergonomic position for the upper-arm is the zero angle, pointing downwards. This is the aim of translation t in Figure 6.1; to move the upper-arm back to the zero-angle position, i.e. the ergonomic optimum.

The signs for the final values of the translation are adjusted based on the defined axes directions, shown in the upper right corner of Figure 6.1. Note that if $α_s$ is negative, meaning the arm is deviated towards the human user's back, the same equations still hold for the translation value, as the direction of the translation in the y-axis is adjusted through the negative value of $α_s$ negating the overall value for the y translation.

For deviations from the ergonomic zone in the coronal plane, i.e. if the upper-arm is abducted, the response is similar, but applied in different axes, refer to Figure 6.2. The upper-arm being abducted, in the case of a right-handed user, means they are reaching further out to the right. Thus, to correct the posture, the robot must move the workpiece to the user's left, forcing them to bring their upper-arm back into the ergonomic zone. The calculations for this are shown in Figure 6.2.

Here, again, 'a' is the length of the upper-arm. However, the arm is being viewed from behind the human – note the change in axes directions on the upper right corner of Figure 6.2 as compared to Figure 6.1. The angle $α_c$, is therefore showing the human's upper-arm deviated to the right. Translation t aims to bring the upper-arm's end point to the left and down, back to its ergonomic optimum. Calculations are similar to that of Figure 6.1. The final translation value signs are again adjusted according to the defined axes, and also hold for left-handed persons.

The above calculations represent all translations needed to improve the ergonomics of the upper-arm. By making the translations dependent on the monitored angles, all different types of non-ergonomic deviations for the shoulder joint are covered by two main translations. The shoulder-elbow length 'a' does not need to be pre-programmed, as the Kinect™ system provides position points for both the shoulder and the elbow, and therefore, the value for 'a' can be calculated directly from the Kinect™ data, as the magnitude of the shoulder-elbow vector. Including the value 'a' in the calculations enables the protocol to automatically adjust to each individual user.

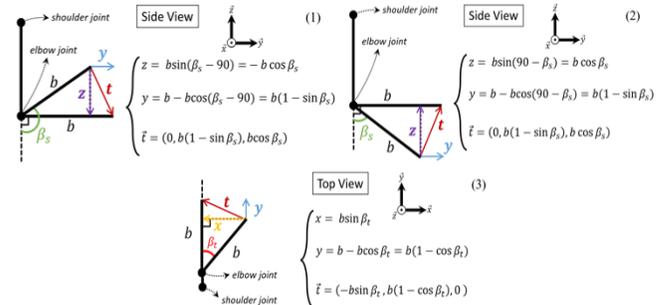

**Figure 7** – (1) Calculations for the translation to improve the lower-arm's deviation in the *sagittal* plane, for angles larger than $100^0$. (2) Calculations for the translation to improve the lower-arm's deviation in the *sagittal* plane, for angles smaller than $60^0$. (3) Calculations for the translation to improve the lower-arm's deviation in the transversal plane. Value 'b' represents the user's elbow-wrist length.

- *Lower-arm responses*

Considering the RULA directives for the lower-arm (Figure 7) there are clear values for different angle ranges for deviation within the sagittal plane. If the lower-arm is deviated to the right or left, going outside the sagittal plane and showing angles projected on the transversal plane, an extra point is added to its overall score. The latter can be detected through the transversal angle described earlier in this paper (refer to Figure 5). If the transversal deviation angle is larger than the threshold of 10⁰, the lower-arm will be considered to be outside the sagittal plane and in transversal deviation. This threshold is chosen so as to avoid unwanted reactions for small deviations.

Similar to calculations for the upper-arm responses, to respond to deviations of the lower-arm in the sagittal plane, an ergonomic optimum target area needs to be considered, so that calculations are based on reaching that optimum.

The ergonomic optimum area for the lower-arm, based on Figure 7, is within the 60⁰-100⁰ range. The value 90⁰ will be considered as the ergonomic target area, which was also considered by the creators of RULA as the optimum point [16].

Therefore, the non-ergonomic cases will be if the sagittal angle is higher than 100⁰ (case 1, Figure 7.1) or lower than 60⁰ (case 2, Figure 7.2) and in both cases the goal is to bring the lower-arm into the 90⁰ position. Calculations for case 1, similar to the upper-arm's sagittal deviations, are shown in Figure 7.1.

Calculations for case 2 are shown in Figure 7.2. As can be seen, the result of calculations, considering axis directions, end up the same as for case 1, i.e.,

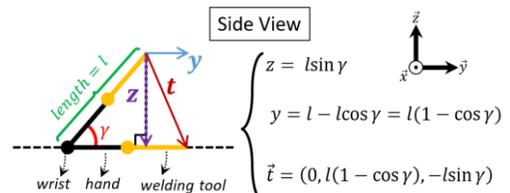

**Figure 8** – Calculations for the translation to improve the wrist's deviation in the *sagittal* plane. (See also Figure 2 (Step 3)).

$$\vec{t} = (0, b(1 - \sin\beta_s), b\cos\beta_s)$$

Deviations of the lower-arm in the transversal plane are shown in the calculations of Figure 7.3. In this case, the robot will need to move the workpiece away from the human, and to their left. This is achieved through the translation described in Figure 7.3. This response was calculated for deviations to the human user's right. Left hand motions use the same equations.

- *Wrist responses*

The wrist responses are implemented slightly differently to the responses for the upper-arm and lower-arm. Without loss of generality we do not consider twisting movements of the wrist, as not needed in a task like a welding operation. Two types of deviations remain: (i) deviations in the transversal plane and (ii) deviations in the sagittal plane. Figure 8 shows the RULA indications for the wrist deviations. Those marked as '+2' and '+3' are transversal, because of the way the wrist is held during welding, i.e. with the palm parallel to the sagittal plane. The case 'Add +1' is a sagittal plane deviation.

When the wrist's transversal angle is in the non-ergonomic range, it is because the workpiece orientation is suboptimal. The response to deviations in the transverse plane will therefore be a rotation of the workpiece rather than a translation.

In case (i) described above, considering there is no wrist twist, and that the only deviation is within the transversal plane, the robot needs only to rotate the workpiece by the same angle and in the opposite direction.

For case (ii), where the deviation is in the sagittal plane, i.e. the wrist is bent upwards or downwards, the correct response will again be a translation, as the workpiece is either too high or too low and needs to be moved into a better position. The response and its calculations are shown in Figure 8.

It must be noted that within these calculations, the tool length (such as a welding tool) needs to be considered as well, in order to achieve the correct translation. Average tool and hand size are used and pre-programmed into the algorithm, as they cannot be measured through the Kinect™. The above calculations hold for up (positive angle) and down (negative angle) movements.

- *Calibration of the Posture Sensors*

While the Kinect™ and Inertial Measurement Unit (IMU) sensors provide absolute values, there are some offsets observed in certain cases, particularly due to the difference in the users' anatomy, which can affect most of all the wrist angle as the IMUs are placed on the arm and their orientation will depend on muscle shapes, but also sometimes the Kinect™ calculated angles. To avoid errors, an initial calibration step is performed for each participant.

This consists of the person, standing in what is defined as the calibration posture, which is based on the ergonomic optimum: The upper-arm at the $0°$ position, lower-arm at $90°$, without any coronal or transversal deviations. The palm is to be held parallel to the sagittal plane, with $0°$ wrist angle. Once the participant is in this position, all joint angles are adjusted to these references. For the wrist angle, the orientation matrix converting the orientation of IMU2 (the one placed on the wrist) to that of IMU1 (the one placed on the back of the hand – refer to Figure 3) is calculated and considered as reference, so that all orientation changes observed in IMU1 are calculated with reference to the original calibrated orientation difference between the two IMUs. In this manner, the angle of the wrist will be calculated independent of the initial deviation between the two IMUs which is due to the user's anatomy. The shoulder-elbow length ('a' in Figure 6) and the elbow-wrist length ('b' in Figure 7) are also recorded, defining the elbow-hip and wrist-elbow distances needed for the flags of §III.A.

C. *Active robot assistance for enhanced ergonomics*

Our approach generates responses for low-ergonomic postures in different parts of the arm, in the framework of RULA, to bring the collaborating human user back into an ergonomically optimum posture. The base of our algorithm is the RULA arm score indicating of whether there is a need for ergonomic improvements. Once activated, our algorithm will go through a series of arm section-specific RULA score checks, to identify the source of the ergonomic deviation. For each identified deviation, the correction as defined in §III.B, is realized by the robot. Our algorithms checks continuously for ergonomic issues and applies improvements where necessary.

Based on the RULA scores providing angles and estimations of which areas around the body would be considered ergonomic, or non-ergonomic, our algorithm considers small virtual windows in front of the human that would be ergonomically optimum to work in. The implementation of our algorithm starts by checking the RULA score for the arm – if the score is 1, meaning totally ergonomic, the algorithm does not react but simply keeps monitoring the score for changes. A RULA arm score of 2 or higher indicates that the human is leaving the ergonomic window. The value selected here as 2, results in a very strict ergonomic window, following RULA very closely. This value can however be adjusted. A higher threshold will lead to less assistive responses by the robot.

If a RULA arm score larger than 1 is detected, it is necessary to assist the human, but first, the cause of the problem is to be identified and a prioritisation is needed on which cause to attend to first. The priority with which different arm sections are checked for problems and responded to is based on which of these sections is the biggest cause of the human leaving the ergonomic window. Highest priority is assigned to deviations within the coronal and transversal planes for the upper and lower-arm, as they happen frequently if the human is trying to reach further, and lead to large increases on the RULA score.

The upper-arm and lower-arm sagittal deviations have a breathing room before leaving the ergonomic window (lower than $20°$ for the upper-arm and between $60°$-$100°$ for the lower-arm are considered ergonomic) and do not immediately cause ergonomic problems. But this is not the case for the wrist, as any deviation will quickly lead to a rise in the ergonomics score. Therefore, the next priority to be checked and responded to is the wrist. The last priority is assigned to deviations of the upper-arm and lower-arm in the sagittal plane. Here, the upper-arm is given priority, as when the upper-arm and lower-arm are both in sagittal deviation, fixing the upper-arm typically leads the human user to fix the lower-arm issue intuitively.

### III. EXPERIMENTAL STUDY AND EVALUATION

To evaluate the ergonomic improvement of the new approach, an experiment is designed. To run the algorithm on an actual robot and perform tests and experiments, the Baxter® Research Robot from Rethink Robotics™ was used. A single arm of the Baxter is used; the other arm is idle. The Baxter screen is used to show the RULA arm score in real-time, with colour coded title cards for the user's information.

The experiment consists of performing a set of targeted movements, while RULA score and Electromyography signals from the participant are recorded. The latter is taken from the following muscles: Trapezius, Deltoid, Triceps Brachii, Biceps Brachii, Wrist Extensors and Wrist Flexors to be used with the Muscle Effort Score (MES) created by the authors in [7]. The pre-planned task is to be performed in two modes: (1) without robot assistance ('human-only') and (2) with robot assistance ('robot-assisted'), to evaluate whether or not our algorithm improves the ergonomics of the collaborative task.

The experimental setup consists of a cardboard box, held by the Baxter robot, representing the workpiece. The participant is directed to reach to certain target areas on the workpiece, while holding a load of 900g in their hand. This load is chosen to simulate a typical welding gun's weight. The target areas are marked and numbered clearly on the workpiece, and shown to the participant, to make sure all participants go through the same procedure. The standing position for all participants is the same and is marked on the floor for consistency. The participants are asked not to move their feet nor use any movements of the trunk and back to reach the target areas, but rather to rely only on the movements of the arm. The trajectory and its timing are designed to last under 3 minutes. Each participant is to go through the trajectory 3 times in each mode, thus a total of 6 trials per person. Throughout the trials, the participant's muscle activity is recorded, using a custom-made wearable setup, described in further detail in [7]. The RULA score is recorded in real-time using the Kinect™ setup (refer to §III). A calibration step is conducted at the beginning of each trial (see §III.B).

5 participants took part in these experiments with an age range of 26.5±2.5 and height range of 179±11cm. For the robot-assisted mode of the test, each participant is asked to reach for the target areas in sequence (refer to Figure 9 (left)); if this leads to a non-ergonomic posture, the robot will respond by moving the workpiece to what it considers to be a more ergonomic position. The participant will follow the workpiece and keep on target. Once the robot's response is complete, the participant is asked to stay on target for 20seconds. This is repeated for all 6 target areas while real-time EMG and RULA for the 20seconds of targeting are recorded. The same process is performed in human-only mode. In this mode, the robot is only keeping the workpiece steady in its predefined initial position. The participant will reach to the target area and keep on target for 20seconds, similar to the robot-assisted approach, however without any movement of the workpiece by the robot. EMG and RULA are again recorded for these periods.

EMG results are investigated using the Muscle Effort Score (MES) method created by the author [7]. Final MES results show no significant difference between the robot-assisted and human-only approaches in terms of external load. This is to be expected, as the MES method provides insight on the external load being handled by the participant, and this is the same in both cases for this experiment. Also, for the performed tasks, the postures do not reach beyond the medium range of RULA, thus no significant intrinsic muscle loading effects are present.

RULA results show a clear difference between the two modes of operation. Figure 9 (right) shows RULA values averaged across participants for each target area (targets 1 to 6, Figure 9 (left)) for both robot-assisted and human-only modes. The robot-assisted mode keeps a steady RULA value around 1, which is the lowest score. Values higher than 1 are observed in cases where the participants are close to the edge of the ergonomic window, thus the overall RULA value is changing between 1 and 2 with small movements, leading to an average higher than 1. The robot does not respond as these changes are happening in under a second considered transient. For the human-only mode, RULA results change for different target areas between high and low values. Since even-numbered targets are already at a comfortable reaching position, these targets are showing values close to '1', also in the human-only results; odd-numbered targets are showing higher RULA values of up to '3'. In contrast, the robot-assisted method is consistently in the ergonomic range, Figure 9 (right).

### IV. CONCLUSIONS

The main aim of our work is to create an ergonomics-based approach that employs robot assistance to improve human posture in the work environment. This new approach is successful in achieving real-time robot-aided posture improvement during manual work, making use of the well-established and widely-used posture ergonomics assessment method, RULA. The new approach has been tested using a Baxter Robot in a collaborative scenario. Results show that enabling the proposed robot-assisted ergonomic interaction approach is consistently improving ergonomics for all users that took part in the study.

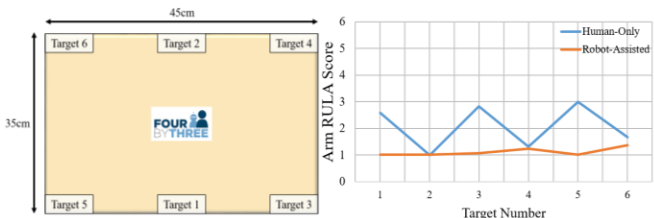

**Figure 9** – (left) Placement of targets on the cardboard workpiece. Odd numbers are on the bottom side and even numbers on the top side. Participants will reach for these targets in sequence; (right) RULA results for the human-only and robot-assisted experiments averaged across all participants and trials. Results show a significant improvement on ergonomics with robot assistance.

## REFERENCES

[1] I. Maurtua *et al.*, "FourByThree: Imagine humans and robots working hand in hand," in *Emerging Technologies and Factory Automation (ETFA), 2016 IEEE 21st International Conference on*, 2016, pp. 1–8.

[2] V. a. Kulyukin and C. Gharpure, "Ergonomics-for-one in a robotic shopping cart for the blind," *Proceeding 1st ACM SIGCHI/SIGART Conf. Human-robot Interact. - HRI '06*, no. November, p. 142, 2006.

[3] A. Schiele and F. C. T. Van Der Helm, "Kinematic design to improve ergonomics in human machine interaction," *IEEE Trans. Neural Syst. Rehabil. Eng.*, vol. 14, no. 4, pp. 456–469, 2006.

[4] F. Zacharias, C. Schlette, F. Schmidt, C. Borst, J. Rossmann, and G. Hirzinger, "Making planned paths look more human-like in humanoid robot manipulation planning," *Proc. - IEEE Int. Conf. Robot. Autom.*, pp. 1192–1198, 2011.

[5] E. S. Boy, E. Burdet, C. L. Teo, and J. E. Colgate, "Investigation of motion guidance with scooter cobot and collaborative learning," *IEEE Trans. Robot.*, vol. 23, no. 2, pp. 245–255, 2007.

[6] A. Bannat *et al.*, "Artificial cognition in production systems," *IEEE Trans. Autom. Sci. Eng.*, vol. 8, no. 1, pp. 148–174, 2011.

[7] A. Shafti, "Electromyography-guided and Robot-assisted Ergonomics", Doctoral Thesis, King's College London, 2017.

[8] K. Wansoo, J. Lee, L. Peternel, N. Tsagarakis, and A. Ajoudani, "Anticipatory robot assistance for the prevention of human static joint overloading in human–robot collaboration." *IEEE Robotics and Automation Letters*, vol. 3, no. 1, pp. 68–75, 2018.

[9] A.G. Marin, M.S. Shourijeh, P.E. Galibarov, M. Damsgaard, L. Fritzsch and F. Stulp, "Optimizing Contextual Ergonomics Models in Human-Robot Interaction." *Proc. – IEEE/RSJ Int. Conf. on Intelligent Robots and Systems (IROS)*, pp. 1–9, 2018.

[10] B. Busch, M. Toussaint and M. Lopes, "Planning ergonomic sequences of actions in human-robot interaction." *Proc. - IEEE Int. Conf. Robot. Autom.*, pp. 1916–1923, 2018.

[11] L. Peternel, W. Kim, J. Babič and A. Ajoudani, "Towards ergonomic control of human-robot co-manipulation and handover." *Proc. - IEEE Int. Conf. on Humanoid Robotics*, pp. 55–60, 2017.

[12] H. Haggag, M. Hossny, S. Nahavandi, and D. Creighton, "Real time ergonomic assessment for assembly operations using kinect," *Proc. - UKSim 15th Int. Conf. Comput. Model. Simulation, UKSim 2013*, pp. 495–500, 2013.

[13] J. A. Diego-Mas and J. Alcaide-Marzal, "Using Kinect$^{TM}$ sensor in observational methods for assessing postures at work," *Appl. Ergon.*, vol. 45, no. 4, pp. 976–985, 2014.

[14] C. C. Martin *et al.*, "A real-time ergonomic monitoring system using the Microsoft Kinect," *2012 IEEE Syst. Inf. Eng. Des. Symp. SIEDS 2012*, pp. 50–55, 2012.

[15] J. Han, L. Shao, D. Xu, and J. Shotton, "Enhanced computer vision with microsoft kinect sensor: A review," *IEEE Trans. Cybern.*, vol. 43, no. 5, pp. 1318–1334, 2013.

[16] L. Mcatamney and E. N. Corlett, "RULA: a survey method for the investigation of world-related upper limb disorders," *Appl. Ergon.*, vol. 24, no. 2, pp. 91–99, 1993.